%% file: main.tex
\pdfoutput=1
\documentclass[11pt]{article}

\usepackage{fairc_arxiv}
\usepackage{xspace}
\usepackage{placeins}
\usepackage{tikz}
\usetikzlibrary{arrows.meta,positioning}

\title{OpenLanguageModel: Readable and Composable\\
Small-Language-Model Pretraining for Education and Research}
\input{author-info}
\input{metadata}

\begin{document}
\maketitle

\begin{abstract}
\input{sections/00-abstract}
\end{abstract}

\input{sections/01-introduction}
\input{sections/04-demonstration}
\input{sections/03-system-design}
\input{sections/05-evaluation}
\input{sections/02-positioning}
\input{sections/07-conclusion}

\FloatBarrier

\section*{Limitations}
\input{sections/08-limitations}

\section*{Ethical Considerations}
\input{sections/09-ethics}

\section*{Acknowledgements}
\input{sections/11-acknowledgements}

\clearpage
\begingroup
\small
\setlength{\bibsep}{3pt}
\bibliographystyle{plainnat}
\bibliography{references}
\endgroup

\clearpage
\appendix
\input{sections/10-appendix}

\end{document}

%% file: author-info.tex
\author{%
  \begin{tabularx}{0.92\linewidth}{@{}>{\centering\arraybackslash}X>{\centering\arraybackslash}X@{}}
    \person{Tavish Mankash}{tavish@mankash.com} &
    \person{Vardhaman Kalloli}{vardhamankalloli722@gmail.com}\\[0.8em]
    \person{Keshava Prasad}{keshava.prasad07@gmail.com} &
    \person{Deepan Muthirayan}{deepan.muthirayan@plaksha.edu.in}
  \end{tabularx}
}

%% file: metadata.tex
\newcommand{\ProjectURL}{https://openlanguagemodel.github.io/openlanguagemodel/}
\newcommand{\RepositoryURL}{https://github.com/openlanguagemodel/openlanguagemodel}
\newcommand{\PackageURL}{https://pypi.org/project/openlanguagemodel/}
\newcommand{\DemoVideoURL}{https://youtu.be/7TdhK-gw4gE}

%% file: sections/00-abstract.tex
OpenLanguageModel (OLM) is an open-source PyTorch library for building and pretraining small language models while keeping their machinery visible. In OLM, model code reads like the architecture: components are ordinary modules, while \texttt{Block}, \texttt{Residual}, \texttt{Repeat}, and \texttt{Parallel} describe how they are wired. The resulting model can move unchanged from a teaching notebook to a complete pretraining run or a research ablation. OLM connects this readable model layer to tokenizers, local and streaming datasets, optimization, mixed precision, callbacks, checkpoints, and hardware-aware CPU, single-GPU, and single-node multi-GPU execution. We demonstrate the full path by tracing GPT-2 from diagram to code, launching a FineWeb-Edu training script, replacing one attention component, and letting \texttt{AutoTrainer} configure the available machine. The package includes 27 presets across nine familiar model families and documentation that progresses from LM fundamentals to architecture research. Validation shows close agreement with independent reference implementations, 90.6\% four-GPU weak-scaling efficiency for a 348M-parameter workload, compact architecture edits, and positive early usability results. OLM is MIT-licensed and available through PyPI, GitHub, and its documentation site.

%% file: sections/01-introduction.tex
\section{Introduction}
\label{sec:introduction}

Small language models (SLMs) put the complete language-modelling workflow within reach of a classroom, a single workstation, or a focused research project. They are large enough to exercise real tokenization, data streaming, optimization, checkpointing, and distributed training, yet small enough that one person can still understand and change the whole model.

OpenLanguageModel (OLM) brings those two goals---understanding the model and running it seriously---into the same PyTorch library. Its model definitions follow the diagrams used to explain transformers. Embeddings, attention, normalization, feed-forwards, residual paths, repeated layers, and branches remain visible as ordinary modules. The same object then connects to a complete SLM pretraining stack, carrying a lesson directly into an experiment.

This design serves students and educators who want to build an LM from first principles, practitioners who want a readable pretraining recipe, and researchers who want local component experiments with the surrounding system held steady. Users can start from a named reference model, assemble a model from components, insert a handwritten \texttt{nn.Module}, or bring the resulting model to their own PyTorch loop.

OLM keeps three things together: architecture code that mirrors the model, end-to-end hardware-aware SLM pretraining, and a learning path that grows into research. The paper follows that workflow directly. We first read a model, train it, change one idea, and scale the run; we then open the system design and give compact validation of correctness, customization, scaling, and usability.

%% file: sections/04-demonstration.tex
\section{A Guided Tour of OLM}
\label{sec:demo}

A first-time user can install OLM with \texttt{pip install openlanguagemodel}. The demonstration then follows one model from source code to a saved checkpoint.

\begin{figure}[t]
  \centering
  \faircfigurebox{%
    \centering
    \includegraphics[width=\linewidth]{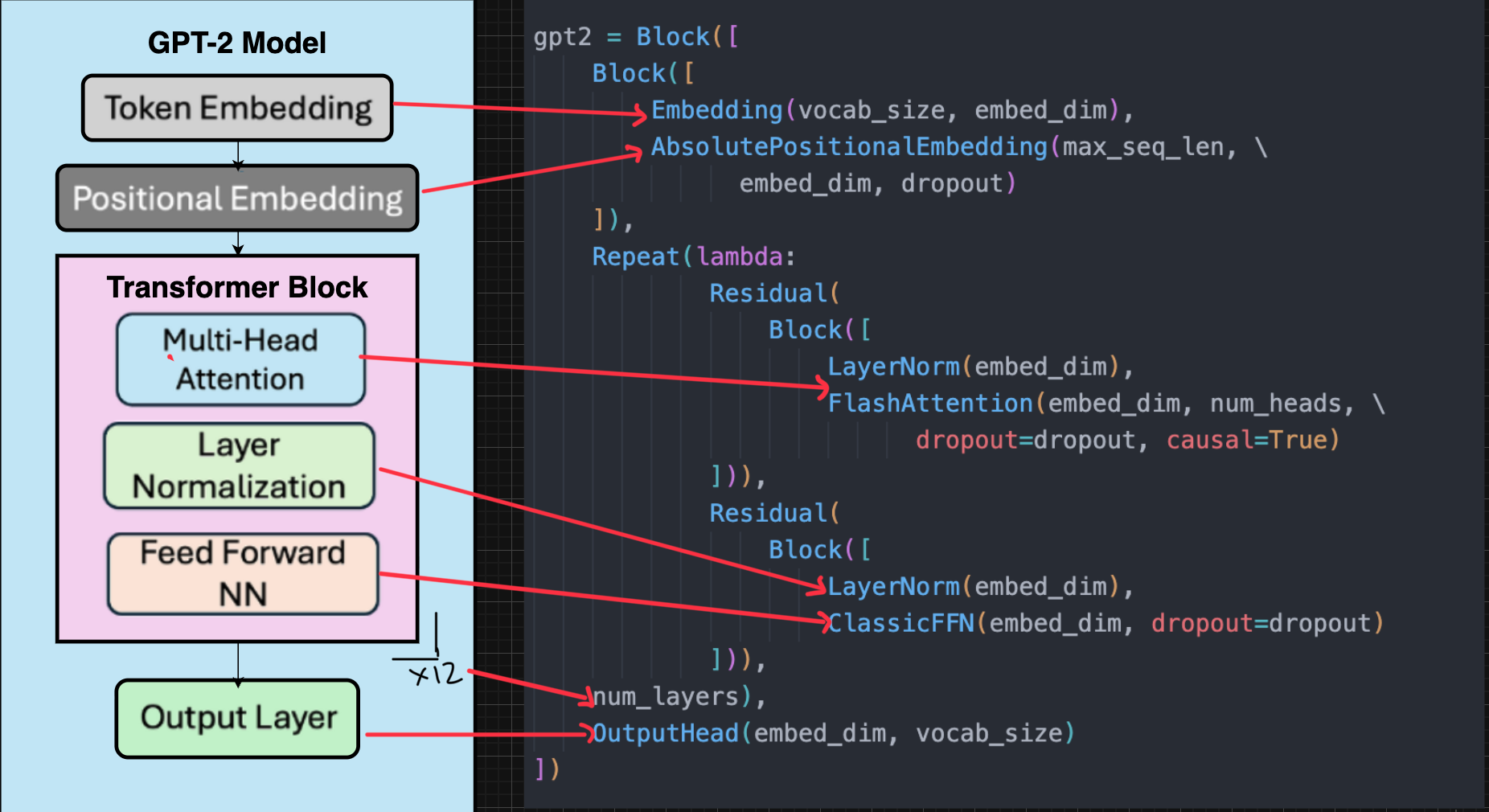}%
  }
  \caption{GPT-2 as a diagram and as an OLM model. The order, nesting, residual paths, and layer repetition are visible in the executable PyTorch definition.}
  \label{fig:gpt2-readable}
\end{figure}

\paragraph{Read the model.}
Figure~\ref{fig:gpt2-readable} places a GPT-2 diagram beside its OLM definition. The correspondence is literal: token and positional embeddings feed a repeated stack of residual attention and feed-forward blocks, followed by an output head. Opening a Llama-style definition reveals the same structural vocabulary with RMSNorm, grouped-query attention, RoPE, and SwiGLU in place of GPT-2's components. The full architecture is visible at a glance in the definition.

\paragraph{Train the same object.}
Figure~\ref{fig:demo-code}A shows a compact pretraining script: choose a tokenizer, stream FineWeb-Edu, construct a familiar model, and pass ordinary PyTorch objects to \texttt{AutoTrainer}. The trainer supplies the training loop, mixed precision, accumulation, scheduling, metrics, and checkpoints. A local text dataset or a user-written loop can slot into the same workflow while the model stays unchanged.

\input{figures/demo-code}

\paragraph{Change one idea.}
The right side of Figure~\ref{fig:demo-code} makes the research workflow concrete. The attention object is one line in an otherwise unchanged transformer layer. Replacing grouped-query attention with ALiBi attention leaves the residual wiring, normalization, feed-forward path, data pipeline, optimizer, and trainer intact. A new attention rule can also subclass \texttt{AttentionBase} or enter the block as any ordinary \texttt{nn.Module}.

\paragraph{Fit the run to the machine.}
With \texttt{device="auto"}, OLM reports the detected devices, available memory, estimated model footprint, and selected execution strategy. The same call runs on CPU or one GPU and configures DDP or FSDP for multiple GPUs on one node. During training, the user sees loss, perplexity, learning rate, and throughput; callbacks can add validation or experiment tracking. The demo ends by saving, reloading, and sampling from the model.

The installable package is at \url{\PackageURL}, and the accompanying demonstration video follows this sequence at \url{\DemoVideoURL}.

%% file: figures/demo-code.tex
\begin{figure}[t]
  \centering
  {\color{faircline}\rule{\linewidth}{0.45pt}}\vspace{0.55em}

  \begin{minipage}[t]{0.485\linewidth}
    \faircpaneltitle{A}{Connect a model to complete pretraining}
    \vspace{2pt}\faircpanelrule\vspace{2pt}
    \scriptsize
\begin{verbatim}
tok = HFTokenizer("gpt2")
data = FineWebEduDataset(
    tok, context_length=1024, streaming=True)
loader = DataLoader(data, batch_size=8)

model = GPT2Model(
    vocab_size=tok.vocab_size,
    embed_dim=768, num_layers=12,
    num_heads=12, max_seq_len=1024)

trainer = AutoTrainer(
    model, AdamW, loader,
    device="auto", context_length=1024,
    learning_rate=3e-4, grad_accum_steps=8)
trainer.train(epochs=1, max_steps=1000)
\end{verbatim}
    \vspace{-4pt}\faircpanelrule
  \end{minipage}
  \hfill
  \begin{minipage}[t]{0.485\linewidth}
    \faircpaneltitle{B}{Make one local architecture change}
    \vspace{2pt}\faircpanelrule\vspace{2pt}
    \scriptsize
\begin{verbatim}
# Baseline component for this run:
attention = GroupedQueryAttention(
    d_model, heads, kv_heads, context)

# For the ALiBi ablation, replace only
# the assignment above with this one:
# attention = MultiHeadAttentionwithALiBi(
#     d_model, heads, max_seq_len=context)

layer = Block([
    Residual(Block([
        RMSNorm(d_model), attention])),
    Residual(Block([
        RMSNorm(d_model), SwiGLUFFN(d_model)])),
])
\end{verbatim}
    \vspace{-4pt}\faircpanelrule
  \end{minipage}

  \vspace{0.55em}{\color{faircline}\rule{\linewidth}{0.45pt}}
  \caption{Two moments from the OLM demo. The left panel connects a readable model to the training stack. The right panel changes the attention rule while preserving the surrounding layer and training path. Imports are omitted for space.}
  \label{fig:demo-code}
\end{figure}

%% file: sections/03-system-design.tex
\section{The System Behind the Demo}
\label{sec:system}

\subsection{Architecture as an Object Graph}
OLM represents both components and wiring with \texttt{torch.nn.Module} objects. \texttt{Block} stores an ordered \texttt{ModuleList}; \texttt{Residual} computes $x+f(x)$; \texttt{Repeat} creates independently parameterized layers from a factory; and \texttt{Parallel} sends one input through multiple branches before a selectable merge. Because the combinators share the same module-in/module-out interface as attention or feed-forward layers, they nest naturally. A parallel residual path can be written as \texttt{Residual(Parallel([attention, ffn]))}; a heterogeneous stack is simply an explicit list of different layers.

The object graph is the model. It can be printed, traversed, placed on a device, optimized, saved, or embedded inside a larger handwritten module using normal PyTorch tools. The trainer consumes this graph directly, so source, inspection, and execution share one representation. Custom control flow can still use a conventional \texttt{forward} method.

\subsection{Reusable Components and Reference Models}
OLM exposes attention, positional encoding, embeddings, normalization, feed-forward layers, output heads, and structural combinators as independent imports. Configurable implementations of \href{https://github.com/openlanguagemodel/openlanguagemodel/blob/main/src/olm/models/openai/gpt2.py}{GPT-2}, \href{https://github.com/openlanguagemodel/openlanguagemodel/blob/main/src/olm/models/meta/llama2.py}{Llama 2}, \href{https://github.com/openlanguagemodel/openlanguagemodel/blob/main/src/olm/models/meta/llama3.py}{Llama 3}, \href{https://github.com/openlanguagemodel/openlanguagemodel/blob/main/src/olm/models/alibaba/qwen2.py}{Qwen 2.5}, \href{https://github.com/openlanguagemodel/openlanguagemodel/blob/main/src/olm/models/microsoft/phi3.py}{Phi-3}, \href{https://github.com/openlanguagemodel/openlanguagemodel/blob/main/src/olm/models/microsoft/phi4.py}{Phi-4}, \href{https://github.com/openlanguagemodel/openlanguagemodel/blob/main/src/olm/models/google/gemma2.py}{Gemma 2}, \href{https://github.com/openlanguagemodel/openlanguagemodel/blob/main/src/olm/models/allenai/olmo.py}{OLMo}, and \href{https://github.com/openlanguagemodel/openlanguagemodel/blob/main/src/olm/models/facebook/opt.py}{OPT} use those same pieces, with 27 named presets for familiar parameterizations. Family blocks and model skeletons remain separate from preset dimensions, making a reference model both a ready starting point and readable example code.

The component layer includes explicit educational implementations as well as efficient paths backed by PyTorch scaled dot-product attention. Modern details such as RoPE, ALiBi, grouped-query attention, RMSNorm, SwiGLU, alternating local attention, and mixture-of-experts feed-forwards remain visible where they enter the model.

\subsection{A Connected Pretraining Stack}
The founding team has trained more than 40 SLMs below one billion parameters across its research projects, and that repeated use has shaped OLM's training path. The library provides local and streaming datasets, Hugging Face tokenizer integration, data loading, optimizers, losses, warmup and decay schedules, mixed precision, gradient accumulation and clipping, callbacks, throughput logging, and checkpoint save/reload.

\texttt{AutoTrainer} inspects CPU and CUDA devices, GPU count and memory, and the estimated model footprint, then chooses the base trainer, DDP, or FSDP according to a balanced, speed, or memory-oriented policy. Users can select a strategy directly or keep the model in their own loop. Architecture and training stay separate, so a component experiment continues to use the same tested data and optimization path.

\subsection{Documentation as Part of the System}
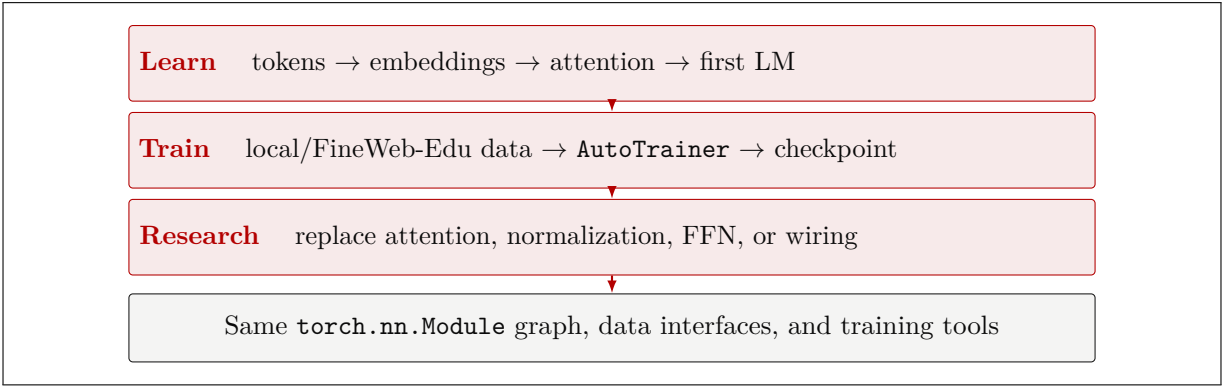
\begin{figure}[t]
  \centering
  \faircfigurebox{\centering\input{figures/architecture-composition}}
  \caption{Three entry points, one code path. Learning, pretraining, and research use the same inspectable modules and connected training stack.}
  \label{fig:architecture}
\end{figure}

The documentation at \url{\ProjectURL} begins with tokens, embeddings, attention, and next-token prediction; continues to a runnable small model with training, generation, and save/reload; and then opens the block system, modern architectures, distributed training, and the API reference. Figure~\ref{fig:architecture} summarizes the three entry points. The code introduced in the beginner path is the same code used later in the research path.

%% file: figures/architecture-composition.tex
\begin{tikzpicture}[
  font=\small,
  >={Latex[length=1.7mm]},
  flow/.style={->, line width=0.8pt, draw=faircred},
  stage/.style={
    draw=faircred,
    fill=fairclightred,
    rounded corners=1.5pt,
    align=left,
    minimum height=10mm,
    text width=125mm,
    inner sep=4pt
  },
  shared/.style={
    draw=faircline,
    fill=fairclightgray,
    rounded corners=1.5pt,
    align=center,
    minimum height=9mm,
    text width=125mm,
    inner sep=4pt
  }
]
  \node[stage] (learn) at (0,0) {
    {\color{faircred}\textbf{Learn}} \quad tokens $\rightarrow$ embeddings $\rightarrow$ attention $\rightarrow$ first LM
  };
  \node[stage] (train) at (0,-1.15) {
    {\color{faircred}\textbf{Train}} \quad local/FineWeb-Edu data $\rightarrow$ \texttt{AutoTrainer} $\rightarrow$ checkpoint
  };
  \node[stage] (research) at (0,-2.30) {
    {\color{faircred}\textbf{Research}} \quad replace attention, normalization, FFN, or wiring
  };
  \node[shared] (shared) at (0,-3.50) {
    Same \texttt{torch.nn.Module} graph, data interfaces, and training tools
  };
  \draw[flow] (learn) -- (train);
  \draw[flow] (train) -- (research);
  \draw[flow] (research) -- (shared);
\end{tikzpicture}

%% file: sections/05-evaluation.tex
\section{Compact System Validation}
\label{sec:evaluation}

The evaluation checks the capabilities shown in the tour: model coverage and reference fidelity, local architecture changes, single-node scaling, and the experience of early users. Table~\ref{tab:evaluation-summary} gives the short version; protocols and packaged records are described in Appendix~\ref{sec:appendix-evaluation}.

\input{tables/evaluation-summary}

\paragraph{Coverage and reference fidelity.}
Reduced configurations of all nine model families passed forward, analytical parameter-count, tied-embedding, and bitwise checkpoint round-trip checks. All 27 named presets passed configuration and parameter-formula checks, and the local suite passed 149 tests. For a numerical check, we copied weights from independent Hugging Face implementations into four-layer FP32 GPT-2, Llama 3, and Qwen 2.5 variants. Across three seeds per family, the maximum absolute logit difference was $3.576\times10^{-7}$, the maximum loss difference was $4.768\times10^{-7}$, and selected gradient cosine similarity remained above $0.9999999999998$.

\paragraph{Local architecture changes.}
We implemented six downstream edits in new user scripts while keeping framework, training, and data files unchanged: parallel residuals, heterogeneous blocks, selected-layer MoE, RoPE-to-ALiBi replacement, cross-family block mixing, and an N-way learned merge. On the four tasks completed in OLM, LitGPT, and Pico, the new scripts contained 117, 171, and 195 lines respectively. Across all six OLM--LitGPT tasks, the totals were 167 and 256 lines. This small code surface reflects reuse of exposed components and combinators.

\paragraph{Single-node scaling.}
A 348,447,744-parameter Llama-3-style model was weak-scaled on one, two, and four A100-SXM4-80GB GPUs with BF16, AdamW, PyTorch SDPA, and 2,048 tokens per GPU. Mean throughput across three runs was 17,785, 32,426, and 64,427 tokens/s. The four-GPU run retained 90.6\% efficiency relative to ideal scaling from one GPU, while peak allocated memory stayed at 10.47 GB per GPU on two and four GPUs.

\paragraph{Early user experience.}
In a de-identified survey of 20 OLM users, the mean System Usability Scale score was 73.8 (95\% bootstrap CI 69.4--78.4). All 20 respondents agreed that the architecture was understandable and that OLM felt natural with ordinary PyTorch; 19 agreed that architecture changes were localized. Among 17 respondents able to compare effort, 14 found architecture-level changes easier in OLM.

%% file: tables/evaluation-summary.tex
\begin{table}[t]
  \centering
  \caption{Validation summary.}
  \small
  \begin{tabularx}{\columnwidth}{@{}>{\raggedright\arraybackslash}p{0.18\columnwidth}>{\raggedright\arraybackslash}p{0.35\columnwidth}>{\raggedright\arraybackslash}X@{}}
    \toprule
    \textbf{Aspect} & \textbf{Setup} & \textbf{Outcome} \\
    \midrule
    Coverage & Reduced models and preset manifests & 9 families; 27 presets; 149 tests \\
    Fidelity & 3 families $\times$ 3 seeds & Max logit diff. $3.58\times10^{-7}$ \\
    Local edits & Four matched tasks & 117 OLM; 171 LitGPT; 195 Pico LoC \\
    Scaling & 348M BF16; 1/2/4 A100s & 64.4k tok/s; 90.6\% at 4 GPUs \\
    User experience & Survey ($n=20$) & SUS 73.8; 20/20 agreed architecture was understandable \\
    \bottomrule
  \end{tabularx}
  \label{tab:evaluation-summary}
\end{table}

%% file: sections/02-positioning.tex
\section{Positioning and Availability}
\label{sec:positioning}

OLM builds on PyTorch's imperative module system~\citep{paszke2019pytorch} and complements several strong language-model ecosystems. Transformers offers broad access to pretrained models~\citep{wolf-etal-2020-transformers}; LitGPT provides compact training recipes~\citep{litgpt-2023}; LMFlow emphasizes foundation-model finetuning and inference~\citep{diao-etal-2024-lmflow}; and Pico supports hypothesis-driven SLM research~\citep{diehl-martinez-etal-2025-pico}. OLM's focus is the path that connects these concerns: readable architecture code, guided learning, complete SLM pretraining, and local research edits in one library.

As of this writing, OLM v2.2.0 is the latest release; the project is actively developed by seven contributors and distributed under the MIT license. The package is installable from \url{\PackageURL}; source is hosted at \url{\RepositoryURL}; and the website at \url{\ProjectURL} provides the beginner course, first-model tutorial, training guides, architecture guide, reference-model source, and generated API documentation. These public entry points make the demonstrated system available for teaching, research, and extension.

%% file: sections/07-conclusion.tex
\section{Conclusion}
\label{sec:conclusion}

OLM starts from a simple promise: the code used to explain a language model should still be useful when it is time to train and change that model. A student can follow the arrows in GPT-2, a practitioner can connect the same object to streaming data and hardware-aware training, and a researcher can replace the attention line while keeping the rest of the run intact.

The demonstration shows that path end to end. OLM combines readable PyTorch composition, a complete SLM pretraining suite, familiar reference models, and documentation that grows with the user. The MIT-licensed package, source, and tutorials are available now for teaching, experiments, and new contributions.

%% file: sections/08-limitations.tex
The validation is scoped to the system shown in the demo. Numerical parity uses reduced models, the GPU study uses synthetic tokens and DDP on one A100 node, and edit size is a proxy for code locality rather than authoring time. Future evaluation will extend to full-scale quality, FSDP, and multi-node training. The 20-person survey is an early convenience sample of recent users, so its scores describe that group rather than the broader LM community.

%% file: sections/09-ethics.tex
OLM is general-purpose research software. Its readable, open-source design supports inspection, while users remain responsible for dataset rights, bias and harmful-output evaluation, access controls, and computational resources. Only de-identified aggregate results from the user study are included in the submission; the row-level workbook remains private.

%% file: sections/11-acknowledgements.tex
Created by members of FAIRC, with contributions from Leap AI Club at Plaksha University.

%% file: sections/10-appendix.tex
\section{Evaluation Details}
\label{sec:appendix-evaluation}

Parity used deterministic FP32 CPU execution and seeds 11, 22, and 33. Scaling used one 2,048-token sequence per GPU, 50 warm-up and 200 measured steps, and three replicates. Edit counts cover new downstream-script lines. SUS uses de-identified aggregates, the standard transform~\citep{brooke1996sus}, and 10,000 bootstrap resamples (seed 20260716).